\documentclass[sigconf]{acmart}
\renewcommand\footnotetextcopyrightpermission[1]{} 
\usepackage{booktabs} %
\usepackage[htt]{hyphenat}

\usepackage{times}  %
\usepackage{helvet}  %
\usepackage{courier}  %
\usepackage{url}  %
\usepackage{graphicx}  %
\definecolor{OliveGreen}{rgb}{0.33,0.41,0.18}
\definecolor{asparagus}{rgb}{0.53, 0.66, 0.42}
\definecolor{britishracinggreen}{rgb}{0.0, 0.26, 0.15}

\renewcommand{\textrightarrow}{$\rightarrow$}

\usepackage[utf8]{inputenc} %
\usepackage[T1]{fontenc}    %
\usepackage{url}            %
\usepackage{booktabs}       %
\usepackage{amsfonts}       %
\usepackage{nicefrac}       %
\usepackage{microtype}      %
\usepackage{verbatim}
\usepackage{siunitx}
\usepackage{algorithm}
\usepackage{algorithmic}
\usepackage{amsmath}
\usepackage{multirow}
\usepackage{amsfonts}
\usepackage{todonotes}
\usepackage{subcaption} %
\usepackage{xcolor}
\usepackage{xspace}
\usepackage{footnote}
\makesavenoteenv{tabular}
\def\xadv{\mathbf{x}_{adv} }
\def\xorig{\mathbf{x}_{orig}}

\def\genattack{\texttt{GenAttack}\xspace}
\def\crossover{\texttt{crossover} }
\def\mutation{\texttt{mutation} }
\newcommand{\bx}{\mathbf{x}}

\newcommand{\sign}{\textnormal{sign}}

\colorlet{kw}{blue}
\definecolor{com}{rgb}{0,0.6,0.3}

\DeclareMathOperator*{\argmax}{arg\,max}

\author[M. Alzantot]{Moustafa Alzantot}
\affiliation{
\institution{UCLA}
\city{Los Angeles, U.S.A}
}
\email{malzantot@ucla.edu}
\author[Y. Sharma]{Yash Sharma}
\affiliation{
\institution{Cooper Union}
\city{New York, U.S.A}
}
\email{sharma2@cooper.edu}

\author[S. Chakraborty]{Supriyo Chakraborty}
\affiliation{
\institution{IBM Research}
\city{New York, U.S.A}
}
\email{supriyo@us.ibm.com}

\author[H. Zhang]{Huan Zhang}
\affiliation{
\institution{UCLA}
\city{Los Angeles, U.S.A}
}
\email{huanzhang@ucla.edu}

\author[C. Hsieh]{Cho-Jui Hsieh}
\affiliation{
\institution{UCLA}
\city{Los Angeles, U.S.A}
}
\email{chohsieh@cs.ucla.edu}

\author[M. Srivastava]{Mani B. Srivastava}
\affiliation{
\institution{UCLA}
\city{Los Angeles, U.S.A}
}
\email{mbs@ucla.edu}

\begin{document}
\title{GenAttack: Practical Black-box Attacks with Gradient-Free Optimization}

 \renewcommand{\shortauthors}{M Alzantot et al.}

\begin{abstract}
Deep neural networks are vulnerable to adversarial examples, even in the black-box setting, where the attacker is restricted solely to query access. Existing black-box approaches to generating adversarial examples typically require a significant number of queries, either for training a substitute network or performing gradient estimation. We introduce GenAttack, a gradient-free optimization technique that uses genetic algorithms for synthesizing adversarial examples in the black-box setting. Our experiments on different datasets (MNIST, CIFAR-10, and ImageNet) show that GenAttack can successfully generate visually imperceptible adversarial examples against state-of-the-art image recognition models with orders of magnitude fewer queries than previous approaches. 
Against MNIST and CIFAR-10 models, GenAttack required
roughly 2,126 and 2,568 times fewer queries respectively, than ZOO, the prior state-of-the-art black-box attack. In order to scale up the attack to large-scale high-dimensional ImageNet models, we perform a series of optimizations that further improve the query efficiency of our attack leading to 237 times fewer queries against the Inception-v3 model than ZOO. Furthermore, we show that GenAttack can successfully attack some state-of-the-art ImageNet defenses, including ensemble adversarial training and non-differentiable or randomized input transformations. Our results suggest that evolutionary algorithms open up a promising area of research into effective  black-box attacks.

\end{abstract}



\maketitle

\section{Introduction}
\label{sec:intro}
Deep neural networks (DNNs) have achieved state-of-the-art performance in various tasks in machine learning and artificial intelligence. Despite their effectiveness, recent studies have illustrated the vulnerability of DNNs to adversarial examples~\cite{szegedy2013intriguing,goodfellow2014explaining}. For instance, a virtually imperceptible perturbation to an image can lead a well-trained DNN to misclassify. Targeted adversarial examples can even cause misclassification to a chosen class. Moreover, researchers have shown that these adversarial examples are still effective in the physical world~\cite{kurakin2016adversarial,athalye2017physical}, and can be crafted in distinct data modalities, such as in the natural language~\cite{alzantot2018generating}, and speech~\cite{alzantot2018did} domains. The lack of robustness exhibited by DNNs to adversarial examples has raised serious concerns for security-critical applications.

Nearly all previous work on adversarial attacks,~\cite{goodfellow2014explaining,carlini2017towards,ead,moosavi2016deepfool,gu2014towards,kurakin2016adversarial} has used gradient-based optimization in order to find successful adversarial examples. However, gradient computation can only be performed when the attacker has full knowledge of the model architecture and weights. Thus, these methods are only applicable in the \textit{white-box} setting, where an attacker is given full access and control over a targeted DNN. However, when attacking real-world systems, one needs to consider the problem of performing adversarial attacks in the \textit{black-box} setting, where nothing is revealed about the network architecture, parameters, or training data. In such a case, the attacker only has access to the input-output pairs of the classifier. A popular approach in this setting have relied on attacking trained substitute networks, and hoping the generated examples transfer to the target model~\cite{papernot2017practical}. This approach suffers from the inherent model mismatch between the substitute model to the target model, as well as the high computational cost required to train the substitute network. Recent works have used coordinate-based finite difference methods in order to directly estimate the gradients from the confidence scores, however the attacks are still computationally expensive, relying on optimization tricks to remain tractable~\cite{chen2017zoo}. Both approaches are query-intensive, thus limiting their practicality in real-world scenarios.

Motivated by the above, we present \genattack, a novel approach to generating adversarial examples without having to compute or even approximate the gradients, enabling efficient adversarial attacks to the black-box case. In order to perform \textit{gradient-free optimization}, we adopt a population-based approach using genetic algorithms, iteratively evolving a population of feasible solutions until success. We also present a number of tricks that allows \genattack to maintain its query-efficiency when attacking models trained on large-scale higher-dimensional datasets, such as ImageNet~\cite{deng2009imagenet}.

Due to its gradient-free nature, \genattack is robust to defenses which perform gradient masking or obfuscation~\cite{athalye2018obfuscated}. Thus, unlike many current black-box attack approaches, \genattack can efficiently craft perturbations in the black-box setting 
to bypass some recently proposed defenses which manipulate the gradients.

We evaluate \genattack using state-of-the-art image classification models, and find that the algorithm is successful at performing \textit{targeted} black-box attacks with significantly less queries than current approaches. In our MNIST, CIFAR-10, and ImageNet experiments, \genattack required roughly $\mathbf{2,126}$, $\mathbf{2,568}$, and $\mathbf{237}$ times less queries than the current state-of-the-art black-box attack, respectively. %
Additionally, we also demonstrate the success of \genattack against state-of-the-art ImageNet defenses, such as ensemble adversarial training~\cite{tramer2017ensemble}, and randomized, non-differentiable input transformation defenses~\cite{facebook}. These results illustrate the power of \genattack's query efficiency and gradient-free nature. 

In summary, we make the following contributions:
\begin{itemize}
\item We introduce \genattack, a novel gradient-free approach for generating adversarial examples by leveraging population-based optimization. Our implementation is open-sourced\footnote{\url{https://github.com/nesl/adversarial_genattack}} to promote further research in studying adversarial robustness.

\item We show that in the restricted black-box setting, \genattack using genetic optimization, as well as dimensionality reduction and adaptive parameter scaling, can generate adversarial examples which force state-of-the-art image classification models, trained on MNIST, CIFAR-10 and ImageNet, to misclassify examples to chosen target labels with significantly less queries than previous approaches.
\item We further highlight the effectiveness of \genattack by illustrating its success against a few state-of-the-art ImageNet defenses, namely ensemble adversarial training and randomized, non-differentiable input transformations. To the best of our knowledge, we are the first to present a successful black-box attack against these defenses.
\end{itemize}

The rest of this paper is organized as follows: Section~\ref{sec:related} provides a summary of related work. Section~\ref{sec:threat} formally defines the threat model for our attack. Section~\ref{sec:algorithm} discusses the details of \genattack. Section~\ref{sec:result} describes our evaluation experiments and their results. Finally, Section~\ref{sec:conc} concludes the paper with a discussion on the utility of evolutionary algorithms for generating adversarial examples.

\section{Related Work}
\label{sec:related}
In what follows, we summarize recent approaches for generating adversarial examples, in both the white-box and black-box cases, as well as defending against adversarial examples. Please refer to the cited works for further details.

\subsection{White-box attacks \& Transferability}

In the \textit{white-box} case, attackers have complete knowledge of and full access to the targeted DNN. In this scenario, the adversary is able to use backpropagation for gradient computation, which enables efficient gradient-based attacks. We briefly summarize a few important white-box attacks formulations below. 

\subsubsection*{L-BFGS}
\cite{szegedy2013intriguing} used box-constrained L-BFGS to minimize the $\ell_2$ norm of the added adversarial noise $\vert\vert \delta \vert \vert_2$ subject to $f(x+\delta)=l$ (prediction is class $l$) and $x+\delta\in[0,1]^m$ (input is within the valid pixel range),

where $f:\mathbb{R}^m\rightarrow\{1,...,k\} $ is the classifier, mapping a data example to a discrete label, $l\in\{1,...,k\}$ is the target output label, and $\delta$ is the added noise.

\subsubsection*{FGSM \& I-FGSM}
\cite{goodfellow2014explaining} proposed the Fast Gradient Sign Method (FGSM) to quickly generate adversarial examples. Under an $L_\infty$ distortion constraint $\| \delta \|_\infty \leq \epsilon$, FGSM uses the sign of the gradient of the training loss $J$ with respect to the original $\bx_0$ and the true label $l$, to generate an adversarial example: $\bx=\bx_0+\epsilon\cdot\sign(\nabla J(\bx_0,l))$, 

Similarly, targeted attacks can be implemented by computing the loss with respect to a specified target class $t$, and instead going in the direction of the negative gradient. 

In~\cite{kurakin2016adversarial}, an iterative version of FGSM was proposed (I-FGSM), where FGSM is used iteratively with a finer distortion constraint, followed by an $\epsilon$-ball clipping. In~\cite{madry}, project gradient descnet (PGD) is introduced, where I-FGSM is modified to incorporate random starts.                   

\subsubsection*{C\&W \& EAD}
Instead of leveraging the training loss, C\&W~\cite{carlini2017towards} designed an $L_2$-regularized loss function based on the logit layer representation in DNNs for crafting adversarial examples. Handling the box constraint $\bx\in[0,1]^p$ using a change of variables, they used Adam~\cite{adam} to minimize $c\cdot f(\bx,t)+\|\bx- \bx_0\|_2^2$, %

where $f(x,t)$ is a loss function depending logit layer values and target class $t$. EAD~\cite{ead} generalizes the attack by minimizing an additional $L_1$ penalty, and has been shown to generate more robust and transferable adversarial examples~\cite{ead_madry,ead_feature,ead_magnet}.

White-box attacks can also be used in black-box settings by taking advantage of \textit{transferability}~\cite{liu2016delving}. Transferability refers to the property that adversarial examples generated using one model are often misclassified by another model~\cite{su2018robustness}. The substitute model approach to black-box attacks takes advantage of this property to generate successful adversarial examples, as we will discuss in the next subsection.

\subsection{Black-box attacks}
In the literature, the \textit{black-box} attack setting has been referred to as the case where an attacker has free access to the input and output of a targeted DNN but is unable to perform back propagation on the network. Proposed approaches have relied on transferability and gradient estimation, and are summarized below. 
\subsubsection*{Substitute Networks}
Early approaches to black-box attacks made use of the power of free query to train a substitute model, a representative substitute of the targeted DNN~\cite{papernot2017practical}. The substitute DNN can then be attacked using any white-box technique, and the generated adversarial examples are used to attack the target DNN. As the substitute model is trained to be representative of a targeted DNN in terms of its classification rules, adversarial examples of the substitute model are expected to be similar to adversarial examples of the corresponding targeted DNN. This approach, however, relies on the transferability property rather than directly attacking the target DNN, which is imperfect and thus limits the strength of the adversary. Furthermore, training a substitute model is computationally expensive and hardly feasible when attacking large models, such as Inception-v3~\cite{szegedy2015going}. 
\subsubsection*{Zeroth Order Optimization (ZOO)}
The zeroth order optimization (ZOO) attack builds on the C\&W attack to perform black-box attacks~\cite{chen2017zoo}, by modifying the loss function such that it only depends on the output of the DNN, and performing optimization with gradient estimates obtained via finite differences. ZOO, however, suffers from requiring a huge number of queries, since a gradient estimate requires $2$ queries per each coordinate. Thus, attacking Inception-v3~\cite{szegedy2015going} on the ImageNet dataset requires $299 \times 299 \times 3 \times 2=536,406$ queries
per each optimization step. To resolve this issue, stochastic coordinate descent (SCD) is used, which only requires 2 queries per step. Still, convergence of SCD can be slow when the number of coordinates is large, thus reducing the dimensionality of the perturbation and using importance sampling are also crucial. Applying these techniques, unlike the substitute model approach, attacking Inception-v3 becomes computationally tractable. However, as we demonstrate in our experiments, the gradient estimation procedure is still quite query-inefficient, and thus impractical for attacking real-world systems.

In parallel works,~\cite{brendel2017decision, ilyas2018black, tu2018autozoom} have also studied the problem of efficiency and strength of black-box adversarial attacks, but our work remains unique in its goal and approach.~\cite{brendel2017decision} focuses on attacking black-box models with only partial access to the query results. Notably, their method takes, on average, about 72x more queries than ours to achieve success against an undefended ImageNet model.~\cite{ilyas2018black} and \cite{tu2018autozoom} study the efficiency of black-box attacks under the same threat model we consider, however, both rely on \textit{gradient estimation}, rather than \textit{gradient-free optimization}.~\cite{ilyas2018black} estimates the gradient of the expected value of the loss under a parameterized search distribution which can be seen as a finite differences estimate on a random gaussian basis.~\cite{tu2018autozoom} dispenses with ZOO's coordinate-wise estimation with a scaled random full gradient estimator. %
Though we treat both efforts as parallel work, for the sake of completeness, we provide a comparison between our ``gradient-free'' approach and the other query-efficient ``gradient-estimation'' approaches in our results.
\subsection{Defending against adversarial attacks}
\subsubsection*{Adversarial Training}
Adversarial training is typically implemented by augmenting the original training dataset with the label-corrected adversarial examples to retrain the network. In~\cite{madry}, a high capacity network is trained against $L_\infty$-constrained PGD, I-FGSM with random starts, yielding strong robustness in the $L_\infty$ ball, but has been shown to be less robust to attacks optimized on other robustness metrics~\cite{ead_madry,schott2018towards,zhang2019limitations}. In~\cite{tramer2017ensemble}, training data is augmented with perturbations transferred from other models, and was demonstrated to have strong robustness to transferred adversarial examples. We demonstrate in our experimental results that its less robust to query-efficient black-box attacks, such as \genattack.

\subsubsection*{Gradient Obfuscation}
It has been identified that many recently proposed defenses provide apparent robustness to strong adversarial attacks by manipulating the gradients to either be nonexistent or incorrect, dependent on test-time randomness, or simply unusable. Specifically, it was found in analyzing the ICLR 2018 non-certified defenses that claim white-box robustness, 7 of 9 relied on this phenomenon~\cite{athalye2018obfuscated}. It has also been shown that FGSM based adversarial training learns to succeed by making the gradients point in the wrong direction~\cite{tramer2017ensemble}. 

One defense which relies upon gradient obfuscation is utilizing non-differentiable input transformations, such as bit-depth reduction, JPEG compression, and total variance minimization~\cite{facebook}. In the \textit{white-box} case, this defense can be successfully attacked with gradient-based techniques by replacing the non-differentiable transformation with the identity function on the backward pass~\cite{athalye2018obfuscated}. Though effective, this approach is not applicable in the \textit{black-box} case, since the attacker requires knowledge of the non-differentiable component. We demonstrate in our experimental results that \genattack, being gradient-free and thus impervious to said gradient manipulation, can naturally handle such procedures in the \textit{black-box} case. Note that many black-box attacks that require gradient estimation including~\cite{chen2017zoo,ilyas2018black, tu2018autozoom} cannot be directly applied when non-differentiable input transformations exist.
\section{Threat Model}
\label{sec:threat}

We consider the following attack scenario. The attacker does not have knowledge about the network architecture, parameters, or training data. The attacker is solely capable of querying the target model as a black-box function:
\[f: \mathbb{R}^d \rightarrow [0,1]^K \]
where $d$ is the number of input features and $K$ is the number of classes. The output of function $f$ is the set of model prediction scores. Note, that the attacker will \textit{not} have access to intermediate values computed in the network hidden layers, including the logits.

The goal of the attacker is to perform a \textit{targeted} attack. Formally speaking, given a benign input example $\mathbf{x}$ that is correctly classified by the model, the attacker seeks to find a perturbed example $\mathbf{x}_{adv}$ for which the network will produce the desired target prediction $t$ chosen by the attacker from the set of labels $\{1..K\}$. Additionally, the attacker also seeks to minimize the $L_p$ distance, in order to maintain the perceptual similarity between $\xorig$ and $\xadv$.
i.e.,

\[
\argmax_{c \in \{1..K\}} f(\mathbf{x}_{adv})
_c = t  \quad
\text{such that }
 \vert\vert \mathbf{x} - \mathbf{x}_{adv} \vert\vert_{p} \leq \delta 
\]
where the distance norm function $L_p$ is often chosen as $L_2$ or $L_{\infty}$.

This threat model is equivalent to that of prior work in black-box attacks~\cite{chen2017zoo,papernot2017practical}, and is similar to the chosen-plain-text attack (CPA) in cryptography, where an attacker provides the victim with any chosen plain-text message and observes its encryption cipher output.

\section{GenAttack Algorithm}
\label{sec:algorithm}
\genattack relies on genetic algorithms, which are population-based gradient-free optimization strategies. Genetic algorithms are inspired by the process of natural selection, iteratively evolving a population $\mathcal{P}$ of candidate solutions towards better solutions. The population in each iteration is called a \textit{generation}. In each generation, the quality of population members is evaluated using a \textit{fitness} function. ``Fitter'' solutions are more likely to be selected for breeding the next generation. The next generation is generated through a combination of \textit{crossover} and \textit{mutation}. Crossover is the process of taking more than one parent solution and producing a child solution from them; it is analogous to reproduction and biological crossover. Mutation applies a small random perturbation to the population members, according to a small user-defined mutation probability. This is done in order to increase the diversity of population members and provide better exploration of the search space. 

Algorithm ~\ref{alg:main} describes the operation of \genattack.
The algorithm input is the original example $\mathbf{x}_{orig}$ and the target classification label $t$ chosen by the attacker. The algorithm computes an adversarial example $\mathbf{x}_{adv}$ such that the model classifies $\mathbf{x}_{adv}$ as $t$ and $|| \xorig - \xadv   ||_{\infty} \leq \delta_{max}$. We define the population size to be $N$, the \textit{``mutation probability''} to be $\rho$, and the \textit{``mutation range''} to be $\alpha$.

\genattack initializes a population of examples around the given input example $\xorig$ by picking random examples from a uniform distributed defined over the sphere centered at the original example $\xorig$, whose radius $=\delta_{max}$. This is achieved by adding random noise in the range $(-\delta_{max},\delta_{max})$ to each dimension of the input vector $\mathbf{x}_{orig}$. Then repeatedly, until a successful example is found, 
each population members' fitness is evaluated, parents are selected, and \crossover \& \mutation are performed to form the next generation. 

\begin{algorithm}[!t]
   \caption{Targeted Adversarial Attack using \genattack}
   \label{alg:main}
\begin{algorithmic}
  \STATE {\bfseries Input:} original example $\xorig$, target label $t$, maximum $L_{\infty}$ distance $\delta_{max}$, mutation-range $\alpha$, mutation probability $\rho$, population size $N$, $\tau$ sampling temperature. 
 \STATE \COMMENT{\textit{Create initial generation}}. 
 \FOR{$i=1,...,N$ in population}
   \STATE $\mathcal{P}^{0}_i \leftarrow  \xorig +  \mathcal{U}(- \delta_{max},\delta_{max}) $
   	\ENDFOR
   	      \FOR{$g=1,2...G \text{ generations}$}
   	       \FOR{$i=1,...,N$ in population}
   	     \STATE $F^{g-1}_i = ComputeFitness(\mathcal{P}^{g-1}_i)$
   	     \ENDFOR
         
          \STATE  \COMMENT{\textit{Find the elite member.}}
   	     \STATE $\xadv = \mathcal{P}^{g-1}_{\argmax_{j} F^{g-1}_j}$
   	     \IF {$\argmax_c f(\xadv)_c == t$ }
   	      \STATE {\bfseries Return: $\xadv$}  \COMMENT{ Found successful attack}
   	     \ENDIF
	\STATE $\mathcal{P}^{g}_1 = \{ \xadv \}$ 
    
 \STATE \COMMENT{\textit{Compute Selection probabilities.}}
	\STATE $probs = Softmax(F^{g-1} / \tau)$ 
	\FOR{$i=2,...,N$ in population}
\STATE $\text{Sample } parent_1 \text{ from }  \mathcal{P}^{g-1} \text{ according to } probs$ 
\STATE $\text{Sample } parent_2 \text{ from }  \mathcal{P}^{g-1} \text{ according to } probs$
\STATE $child = Crossover(parent_1, parent_2) $ 
 \STATE  \COMMENT{\textit{Apply mutations and clipping.}}
		\STATE $child_{mut}$ = $child$ + \parbox[t]{.6\linewidth}{  $Bernoulli(\rho) *  \hookrightarrow$\\
        $\mathcal{U} (-\alpha \, \delta_{max},\alpha\, \delta_{max})$}
        \STATE $child_{mut} = \Pi_{\delta_{max}} (child_{mut}, \xorig)$
         \STATE  \COMMENT{\textit{Add mutated child to next generation}}. 
		\STATE $\mathcal{P}^{g}_i = \{ child_{mut} \}$

	\ENDFOR
   \STATE  \COMMENT{\textit{adaptively update $\alpha$, $\rho$ parameters}}
  \STATE $\rho$, $\alpha$ = UpdateParameters($\rho$, $\alpha$)

   	      \ENDFOR

\end{algorithmic}
\end{algorithm}

\subsubsection*{Fitness function:} The subroutine $\texttt{ComputeFitness}$ evaluates the fitness, i.e. quality, of each population member. As the fitness function should reflect the optimization objective, a reasonable choice would be to use the output score given to the target class label directly. However, we find it more efficient to also jointly motivate the decrease in the probability of other classes. We also find that the use of $\log$ proves to be helpful in avoiding numeric instability issues. Therefore, we chose the following function:
\[ ComputeFitness(\mathbf{x}) = \log{f(\mathbf{x})_t} - \log {\sum_{j=0, j \neq t}^{j=k} {f(\mathbf{x})_c}}\]

\subsubsection*{Selection:} Population members at each iteration are ranked according to their fitness value. Members with higher fitness are more likely to be a part of the next generation while members with lower fitness are more likely to be replaced. We compute the probability of selection for each population member by computing the $\mathbf{Softmax}$ of the fitness values to turn them into a probability distribution. Then, we independently select random parent pairs among the population members according to that distribution. We also find it important to apply the elitism technique~\cite{bhandari1996genetic}, where the \textit{elite} member, the one with highest fitness, of the current generation is guaranteed to become a member of the next generation.

\subsubsection*{Crossover operation:} After selection, parents are mated to produce members of the next generation. A child is generated by selecting each feature value from either $parent_1$ or $parent_2$ according to the selection probabilities $(p, 1-p)$ where $p$ is defined as
\begin{align*}
p = \frac{fitness(parent\_1)}{fitness(parent\_1)+ fitness(parent\_2) }
\end{align*}
\subsubsection*{Mutation operation:} To promote diversity among the population members and exploration of the search space, at the end of each iteration, population members can be subject to mutation, according to probability $\rho$. Random noise uniformly sampled in the range $(-\alpha \, \delta_{\max}, \alpha \, \delta_{\max})$ is applied to individual features of the crossover operation results. Finally, clipping operator $\Pi_{\delta_{max}}$is performed to ensure that the pixel values are within the permissible $L_{\infty}$ distance away from the benign example $\mathbf{x}_{orig}$.

\begin{figure*}[!t]
\begin{minipage}[b]{0.45\linewidth}
\centering
\includegraphics[width=0.9\textwidth]{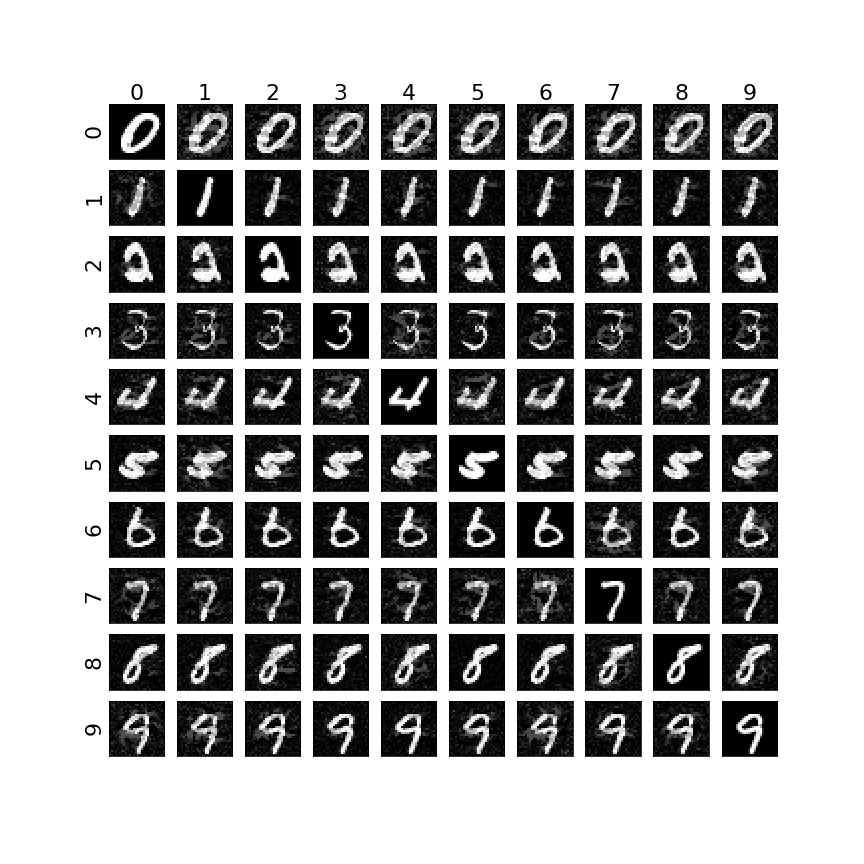}
  \caption{MNIST adversarial examples generated by \genattack. Row label is the true label and column label is the target label.}
  \label{fig:show_mnist}
\end{minipage}
\hspace{0.5cm}
\begin{minipage}[b]{0.45\linewidth}
\centering
\includegraphics[width=0.9\textwidth]{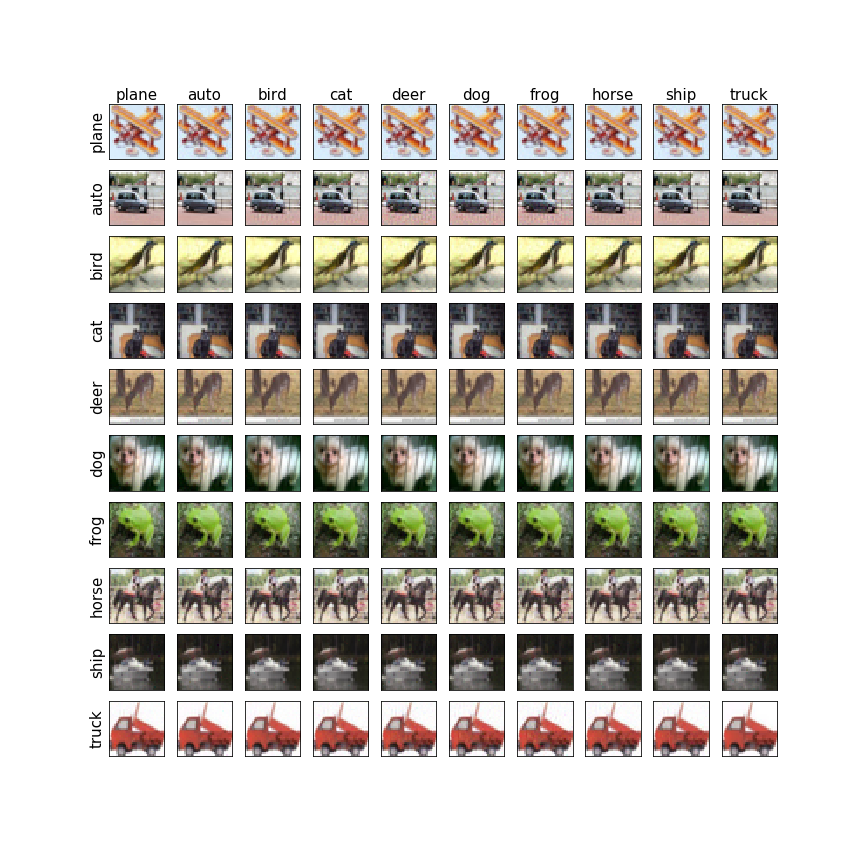}
  \caption{CIFAR-10 adversarial examples generated by \genattack. Row label is the true label and column label is the target label.}  \label{fig:show_cifar}
\end{minipage}
\end{figure*}
\subsection{Improving Query Efficiency}
In this section, we present a couple of optimizations that we adopt in our \genattack algorithm, and which contribute significantly to query efficiency. 
\subsubsection{Dimensionality Reduction:} 
On one hand, scaling genetic algorithms to explore high-dimensional search spaces (such as ImageNet models) efficiently requires a large population size in each generation. On the other hand, evaluating the fitness of each member implies additional costs in the form of new queries. Therefore, we limit \genattack to operate using a relatively small (e.g., less than ten) population size.
We provide more discussion on the trade-off between population size and number of queries in Section ~\ref{sec:hyperparams}. 

In addition, motivated by the work in~\cite{tu2018autozoom}, we seek to address the challenge of scaling genetic algorithms (in turn \genattack), by performing dimensionality reduction of the search space and defining adversarial noise in the lower dimensional space.
To compute the adversarial example, we apply bilinear resizing (which is deterministic), to scale the noise up to same size as the input. Thus,
\begin{align*}
\xorig \in [0, 1] ^ {d}, \quad 
e_{adv} \in [0,1]^{d'}, \quad
\xadv = S(e_{adv}) + \xorig 
\end{align*}
where $e_{adv}$ is the learnt adversarial noise, $S$ is the bilinear resizing operation, and $d'$ is chosen such that $d' < d$. Effectively, this leads to a condensed adversarial noise vector, where one value of $e_{adv}$ is used to perturb multiple neighbouring pixels of $\xorig$ to produce $\xadv$. We noticed that this approach significantly improves the query efficiency of \genattack against high dimensional models, such as ImageNet models, while maintaining the attack success rate under the $L_\infty$ constraint.%

\subsubsection{Adaptive Parameter Scaling}
In order to lessen the sensitivity of genetic algorithms to hyperparameter values (e.g. mutation rate, population size, and mutation range), we use an annealing scheme where the algorithm parameters (namely the mutation rate $\rho$ and mutation range $\alpha$) are decreased gradually if the search algorithm is detected to be stuck for a number of sequential iterations without any further improvement in the objective function. Adaptive scaling alleviates the situation where a very high mutation rate may allow for an initially fast decrease in the objective function value, after which the algorithm may get stuck without achieving any further progress. 

We employ exponential decay to update the parameter values 
\begin{align}
\label{eq:decay_pho}
\rho =  \max(\rho_{\text{min}}, 0.5 \times (0.9)^{\text{num\_plateaus}}) \\
\label{eq:decay_alpha}
\alpha = \max(\alpha_{\text{min}}, 0.4 \times (0.9)^{\text{num\_plateaus}})
\end{align}  
where $\rho_{\text{min}}$ and $\alpha_{\text{min}}$ are chosen to be $0.1$ and $0.15$ respectively, and $\text{num\_plateaus}$ is a counter incremented whenever the algorithm does not improve the fitness of the population's elite member (highest fitness) for $100$ consecutive steps.

\section{Results}
\label{sec:result}
We evaluate \genattack by running experiments attacking state-of-art MNIST, CIFAR-10, and ImageNet image classification models. For each dataset, we use the same models as used in the ZOO work~\cite{chen2017zoo}. For MNIST and CIFAR-10, the model accuracies are 99.5\% and 80\%, respectively. The reader can refer to~\cite{carlini2017towards} for more details on the architecture of those models. For ImageNet, we use Inception-v3~\cite{szegedy2015going}, which achieves 94.4\% top-5 accuracy and 78.8\% top-1 accuracy. We compare the effectiveness of \genattack to ZOO on these models in terms of the attack success rate (ASR), the runtime, and the median number of queries necessary for success. The runtime and query count statistics are computed over successful attacks only. A \textit{single} query means an evaluation of the target model output on a \textit{single} input image. Using the authors' code\footnote{\url{https://github.com/huanzhang12/ZOO-Attack}}, we configure ZOO for each dataset based on the implementations the authors used for generating their experimental results~\cite{chen2017zoo}. We also evaluate against the state-of-the-art \textit{white-box} C\&W attack~\cite{carlini2017towards}, assuming direct access to the model, to give perspective on attack success rate. 

In addition, we evaluate the effectiveness of \genattack against ensemble adversarial training~\cite{tramer2017ensemble}, using models released by the authors at the following link\footnote{\url{https://github.com/tensorflow/models/tree/master/research/adv_imagenet_models}}. Ensemble adversarial training is considered to be the state-of-art ImageNet defense against black-box attacks, proven to be effective at providing robustness against transferred attacks in hosted competitions~\cite{tramer2017ensemble,nipscompetition,caadcompetition}. Finally, we evaluate against recently proposed randomized, non-differentiable input transformation defenses~\cite{facebook} to test \genattack's performance against gradient obfuscation. We find that \genattack can handle such defenses as-is due to its gradient-free nature. 

\subsubsection*{Hyperparameters} 
For all of our MNIST and CIFAR-10 experiments, we limit \genattack to a maximum of 100,000 queries, and fix the hyperparameters to the following values: mutation probability $\rho=\num{5e-2}$, population size $N=6$, and step-size $\alpha=1.0$. For all of our ImageNet experiments, as the images are nearly 100x larger than those of CIFAR-10, we use a maximum of 1,000,000 queries and adaptively update the $\rho$ and $\alpha$ parameters as discussed earlier in Section \ref{sec:algorithm}. In addition, we experimented both with and without dimensionality reduction ($d' = 96$). To match the mean $L_\infty$ distortion computed over successful examples of ZOO, we set $\delta_{max} = \{0.3, 0.05, 0.05\}$, for our MNIST, CIFAR-10, and ImageNet experiments, respectively.
As genetic algorithms have various tuning parameters, we conduct parameter sensitivity studies in Section~\ref{sec:hyperparams}.

\subsection{Query Comparison}
We compare \genattack and ZOO by number of queries necessary to achieve success, and provide C\&W white-box results to put the ASR in perspective. %
For MNIST, CIFAR-10, and ImageNet, we use 1000, 1000, and 100 randomly selected and correctly classified images from the test sets. For each image, we select a random target label.
\subsubsection{MNIST and CIFAR-10:}
Table~\ref{tab:mnist_cifar} shows the results of our experiment. The results show that both ZOO and GenAttack can succeed on the MNIST and CIFAR-10 datasets, however \genattack is 2,126 times and 2,568 times more efficient on each. %
A randomly selected set of MNIST and CIFAR-10 test images and their associated adversarial examples targeted to each other label are shown in Figure~\ref{fig:show_mnist} and Figure~\ref{fig:show_cifar}.

\subsubsection*{ImageNet: } 
\begin{figure*}[ht]
\begin{minipage}[b]{0.45\linewidth}
\centering
\includegraphics[width=0.9\textwidth]{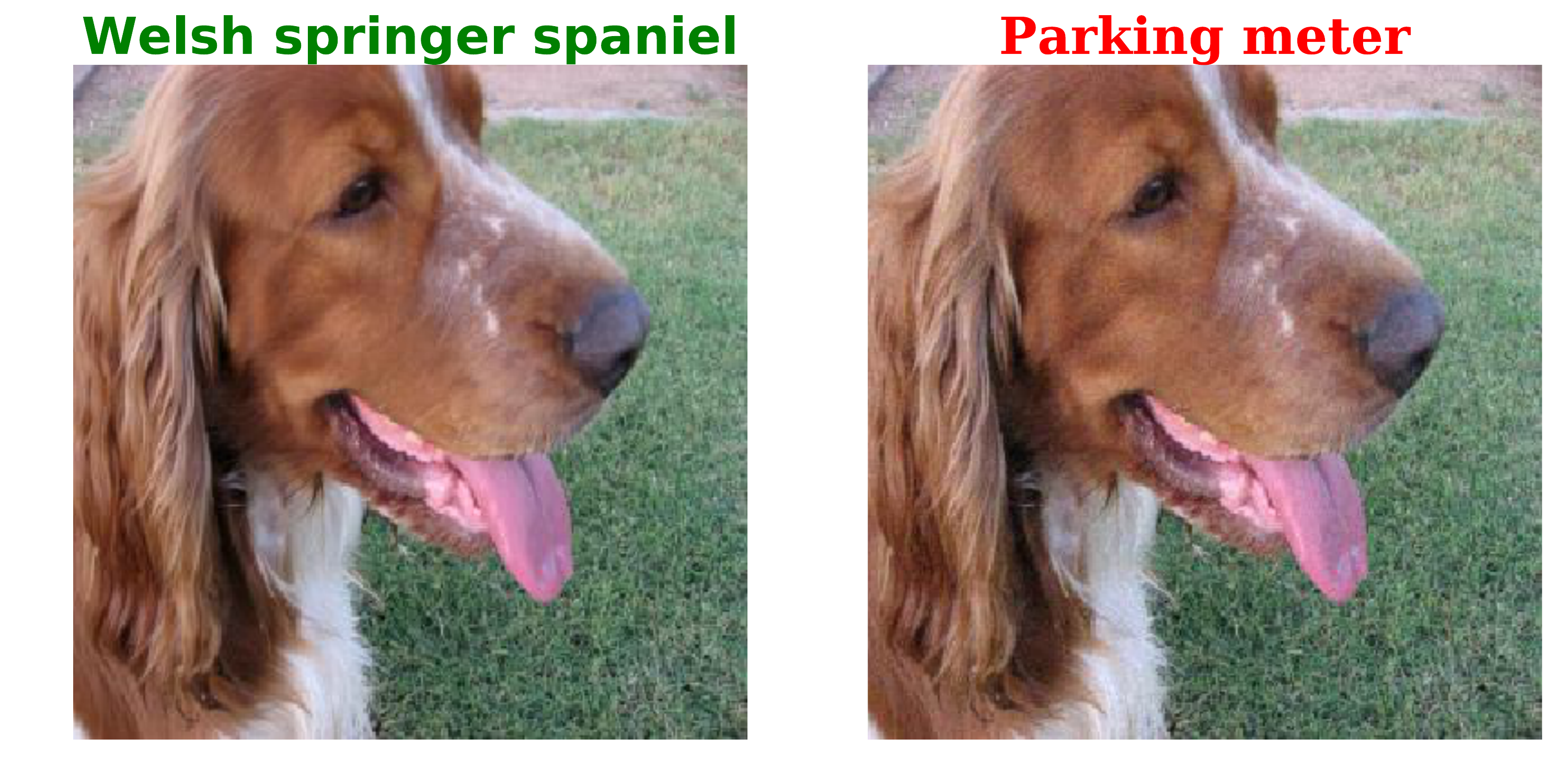}
  \caption{Adversarial example generated by \genattack against the \texttt{InceptionV3} model ($L_{\infty}=0.05$). Left figure: original, right figure: adversarial example.}
  \label{fig:inception:show}
\end{minipage}
\hspace{0.3cm}
\begin{minipage}[b]{0.45\linewidth}
\centering

\includegraphics[width=0.9\textwidth]{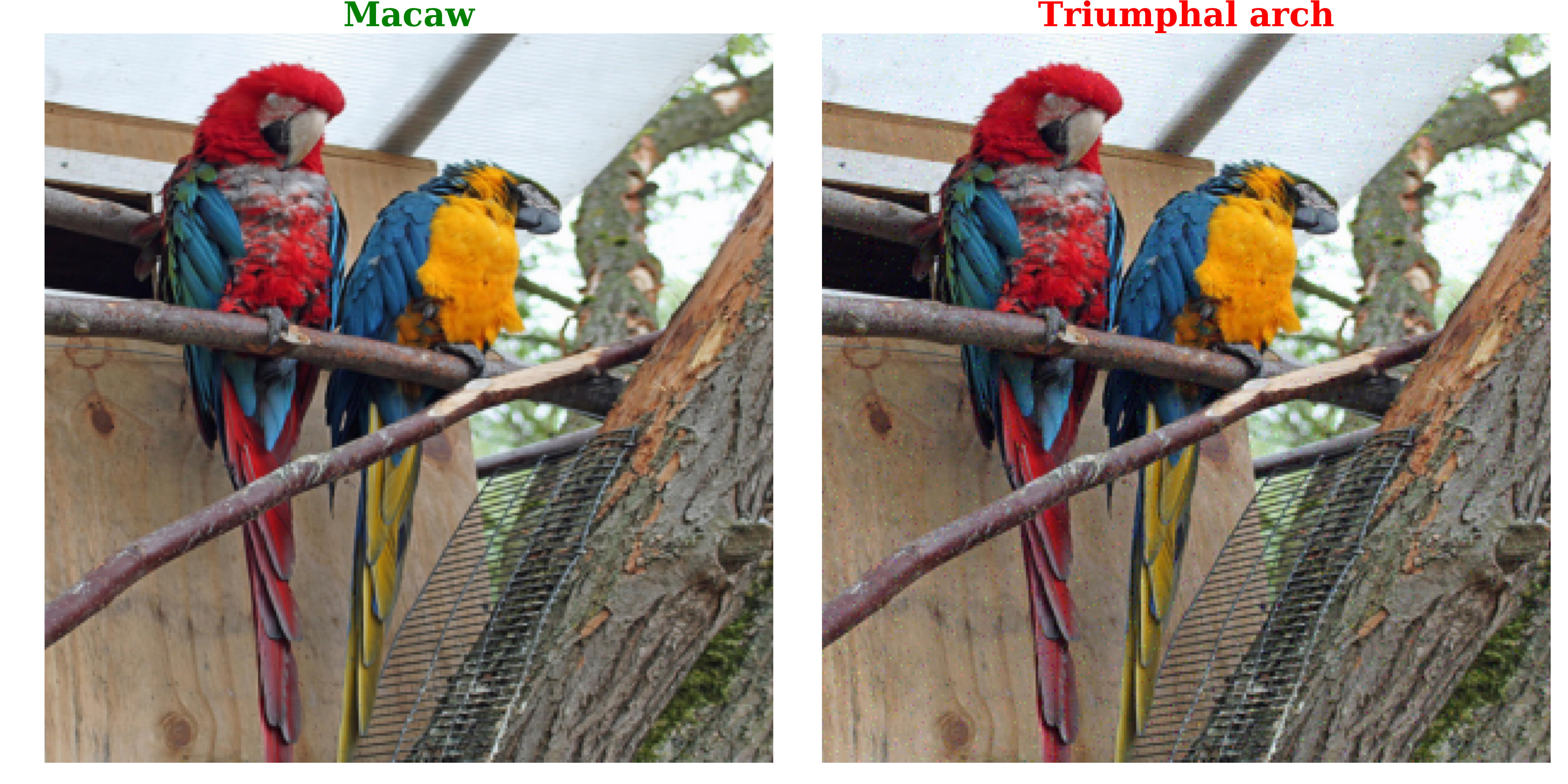}
\caption{Adversarial example generated by \genattack against the \texttt{JPEG} compression defense ($L_{\infty}=0.15$). Left figure: original, right figure: adversarial example.}
\label{fig:advshow}
\end{minipage}
\end{figure*}

Table~\ref{tab:ens_adv} shows the results of our experiment against normally trained (InceptionV3) and ensemble adversarially trained (Ens4AdvInceptionV3) ImageNet models. To demonstrate the effect of dimensionality reduction and adaptive parameter scaling, we denote \genattack without such tricks as ``GA baseline''. On ImageNet, ZOO is not able to succeed consistently in the targeted case, %
which is significant as it shows that \genattack is efficient enough to effectively scale to ImageNet. Moreover, \genattack is roughly $237$ times more efficient than ZOO, and $9$ times more query efficient than the GA baseline. A random example of results against Inception-v3 test image with its associated adversarial example is shown in Figure~\ref{fig:inception:show}. %

\begin{table}[!t]
\centering
\begin{tabular}{c|cc|cc}
\toprule & \multicolumn{2}{c|}{MNIST  \textit{($L_{\infty}=0.30$)}} & \multicolumn{2}{c}{CIFAR-10 \textit{($L_{\infty}=0.05$)} } \\
\hline
 & ASR & Queries  & ASR & Queries \\
\hline
 C\&W  &100\% & -- &   100\% & -- \\
\hline
ZOO &    98\%  &  2,118,222  &   93.3\%  &    2,064,798  \\
\hline
GenAttack &    100\%  & 996  &   96.5\% &  804  \\
\bottomrule
\end{tabular}
\caption{Attack success rate (ASR) and median number of queries for the C\&W (white-box) attack, ZOO, and \genattack with equivalent $L_\infty$ distortion. Median of query counts is computed over successful examples. Number of queries is not a concern for white-box attacks.}
\label{tab:mnist_cifar}
\end{table}

\begin{table}[!h]
\centering
\begin{tabular}{c|cc|cc}
\toprule & \multicolumn{2}{c|}{InceptionV3} & \multicolumn{2}{c}{Ens4AdvInceptionV3} \\
\hline
 & ASR & Queries & ASR & Queries \\
\hline

C\&W & 100\% & - & 100\% & -   \\
\hline
ZOO & 18\% & 2,611,456 & 6\% & 3,584,623 \\
\hline

GA baseline &  100\% & 97,493 &  93\% & 163,995 \\
\hline
GenAttack & 100\% & 11,081 & 95\% & 21,966 \\
\bottomrule
\end{tabular}
\caption{Attack success rate (ASR) and median number of queries for the C\&W (white-box) attack, ZOO, and \genattack with equivalent $L_\infty$ distortion (0.05). Median of query counts is computed over successful examples. GA baseline is \genattack without the dimensionality reduction and adaptive parameter scaling tricks.}
\label{tab:ens_adv}
\end{table}

\subsubsection{Comparison to parallel efforts in query efficient attacks}
While preparing this manuscript, we became aware of parallel efforts that were also developed to address query-efficient adversarial attacks. For sake of completeness, we also present a comparison between our approach and other contributions. Unlike \genattack, which performs gradient-free optimization,~\cite{tu2018autozoom} and~\cite{ilyas2018black} propose more efficient gradient estimation procedures than ZOO. Table~\ref{tab:compare_new} shows a comparison between the results of the three methods. We notice that while the three methods are all significantly more query efficient than the previous state-of-art (ZOO), \genattack requires 25\% fewer queries than the work of~\cite{ilyas2018black} under the same $L_{\infty}$ distance constraint at the cost of a slight increase in $L_2$ distance, mainly due to the use of dimensionality reduction. Also, \genattack requires $15\%$  less queries than~\cite{tu2018autozoom}, even though~\cite{tu2018autozoom} has higher distortion in both $L_{\infty}$ and $L_2$ distortion. Notably, ~\cite{tu2018autozoom} has additional post-processing steps to reduce the amount of distortion but it significantly costs more queries. Therefore, we reported the number of queries and distortion distances at the first success for all attacks.

\begin{table}
    \centering
    \begin{tabular}{c|c|c|c}
    \toprule
    Attack & Queries Count & $L_2$-distance & $L_\infty$-distance \\
    \hline
      \genattack  & \textbf{11,081}  &  $2.3\times 10^{-4}$  & \textbf{0.05}\\
      \hline
       AutoZOOM \cite{tu2018autozoom}  & 13,099 &  $8.1 \times 10^{-4}$ &  0.75  \\
       \hline
       Ilyas et al.\cite{ilyas2018black} & 14,737 &$\mathbf{1.9 \times 10^{-4}}$ & \textbf{0.05} \\
       \bottomrule
    \end{tabular}
    \caption{Comparison with parallel work against the ImageNet InceptionV3 model in terms of both median of query counts and per-pixel-$L_2$ and $L_{\infty}$ distances between the original and adversarial images.}
    \label{tab:compare_new}
\end{table}

\subsection{Attacking Defenses}	
In the following section, we present how \genattack succeeds in breaking a set of state-of-the-art defense methods proposed to increase the robustness of models against adversarial attacks.

\subsubsection{Attacking Ensemble Adversarial Training:}
Ensemble adversarial training incorporates adversarial inputs
generated on other already trained models into the model's training data in order to increase its robustness to adversarial examples~\cite{tramer2017ensemble}. This has proven to be the most effective approach at providing robustness against transfer-based black-box attacks during the NIPS 2017 Competition. We demonstrate that the defense is much less robust against query-efficient black-box attacks, such as \genattack.

We performed an experiment to evaluate the effectiveness of \genattack against the ensemble adversarially trained model released by the authors~\cite{tramer2017ensemble}, namely \texttt{Ens4AdvInceptionV3}. We use the same 100 randomly sampled test images and targets used in our previous ImageNet experiments. We find that \genattack is able to achieve $95\%$ success against this strongly defended model, significantly outperforming ZOO. As seen, in Table~\ref{tab:ens_adv}, we compare the success rate and median query count between the ensemble adversarially trained and the vanilla Inception-v3 models. Our comparison shows that these positive results are yielded with only a limited increase in query count. We additionally note that the maximum $L_\infty$ distortion used for evaluation in the NIPS 2017 competition varied between 4 and 16 (in a 0-255 scale), which when normalized equals 0.02 and 0.06, respectively. Our $\delta_{max}$ (0.05) falls in this range.

\subsubsection{Attacking Non-Differentiable, Randomized Input Transformations:}
Non-differentiable input transformations perform gradient obfuscation, relying upon manipulating the gradients to succeed against gradient-based attackers~\cite{athalye2018obfuscated}. In addition, randomized transformations increase the difficulty for the attacker to guarantee success. One can circumvent such approaches by modifying the core defense module performing the gradient obfuscation, however this is clearly not applicable in the black-box case. 
  \begin{table}[!th]
\centering
\begin{tabular}{c|c|c|c|c|}
\cline{2-5}
& \multicolumn{2}{c|}{CIFAR-10} & \multicolumn{2}{c|}{ImageNet} \\
\cline{2-5}
 & ASR & Queries & ASR & Queries \\
 \hline
 Bit depth & 93\% & 2,796 & 95\% & 16,301 \\
 \hline
 JPEG      & 88\% & 3,541 & 89\% & 23,822  \\
 \hline
 TVM       & 70\% & 5,888 $\times$ 32  & -- & -- \\
\toprule
\end{tabular}
\caption{Evaluation of \genattack against non-differentiable and randomized input transformation defenses. We use $L_{\infty}=0.05$ for bit-depth, and $L_{\infty}=0.15$ for JPEG and TVM experiments. For TVM, we compute the expectation of the fitness function by taking $t=32$ queries.}
\label{tab:facebook}
\end{table}

In~\cite{facebook}, a number of input transformations were explored, including bit-depth reduction, JPEG compression, and total variance minimization. Bit-depth reduction and JPEG compression are non-differentiable, while total variance minimization introduces additional randomization and is quite slow, making it difficult to iterate upon. We demonstrate that \genattack can succeed against these input transformations in the black-box case, due to its \textit{\textbf{gradient-free}} and \textbf{\textit{multi-modal}} population-based nature making it impervious to gradient obfuscation. To the best of our knowledge, we are the first to demonstrate a black-box algorithm which can bypass such defenses. Our results are summarized in Table~\ref{tab:facebook}.

For bit-depth reduction, 3 bits were reduced, while for JPEG compression, the quality level was set to 75, as in~\cite{facebook}. \genattack is able to achieve high success rate against both non-differentiable transformations, on both the CIFAR-10 and ImageNet datasets. A visual example of our results against JPEG compression is shown in Figure~\ref{fig:advshow}.

Total variance minimization (TVM) introduces an additional challenge as it is not only non-differentiable, but it also introduces randomization and is an exceedingly slow procedure. TVM randomly drops many of the pixels (dropout rate of 50\%, as in ~\cite{facebook}) in the original image and reconstructs the input image from the remaining pixels by solving a denoising optimization problem. Due to randomization, the classifier returns a different score at each run for the same input, confusing the attacker. Succeeding against randomization requires more iterations, but iterating upon the defense is difficult due to the slow speed of TVM processing.

In the setting with randomization, the $\texttt{ComputeFitness}$ function can be generalized to be
\[ ComputeFitness(\mathbf{x}) = \mathop{\mathbb{E}}_{r}{[ \log{f(\mathbf{x}, r )_t} - \log {\max_{c \neq t} {f(\mathbf{x},r)_c}} ]}\]
where $f(\mathbf{x}, r)$ is the randomization-defended model query function and $r$ is the noise input to the TVM function, \genattack can still handle this defense in the black-box case. The expectation is computed by querying the model $t$ times (we used $t=32$) for every population member to obtain a robust fitness score at the cost of an increased number of queries. Due to the computational complexity of applying TVM on each query, we performed the TVM experiment only using the CIFAR-10 dataset and achieved  70\% success with $L_\infty$ = $0.15$. Due to the large randomization introduced by TVM, we counted an adversarial example as success only if it is classified as the target label three times in a row. Notably, TVM significantly decreases the model accuracy on clean inputs (e.g. in our CIFAR-10 experiments, from 80\% to 40\%) unless the model is re-trained with transformed examples~\cite{facebook}.

\subsubsection*{Comparison to ZOO and C\&W:}
Due to the non-differentiable nature of the input transformations, the C\&W attack, a gradient-based attack, can not succeed without manipulating the non-differentiable component, as discussed in~\cite{athalye2018obfuscated}. In the \textit{white-box} case, this method can be applied to yield high success rate, but is not applicable in the more restricted \textit{black-box} case. In the black-box setting, ZOO achieved 8\% and 0\% against the non-differentiable bit-depth reduction and JPEG compression defenses on ImageNet, again demonstrating its impracticality.

\subsection{Hyper-parameters values selection}
\label{sec:hyperparams}
Since genetic algorithms are traditionally sensitive to the choice of hyper-parameter values (e.g. population, mutation rate, etc.), we present a discussion regarding this effect, in the context of query efficiency, which leads to our selection of the hyper-parameter values listed in Section~\ref{sec:result}.
\begin{figure}[!t]
    \centering
    \includegraphics[width=0.9\columnwidth]{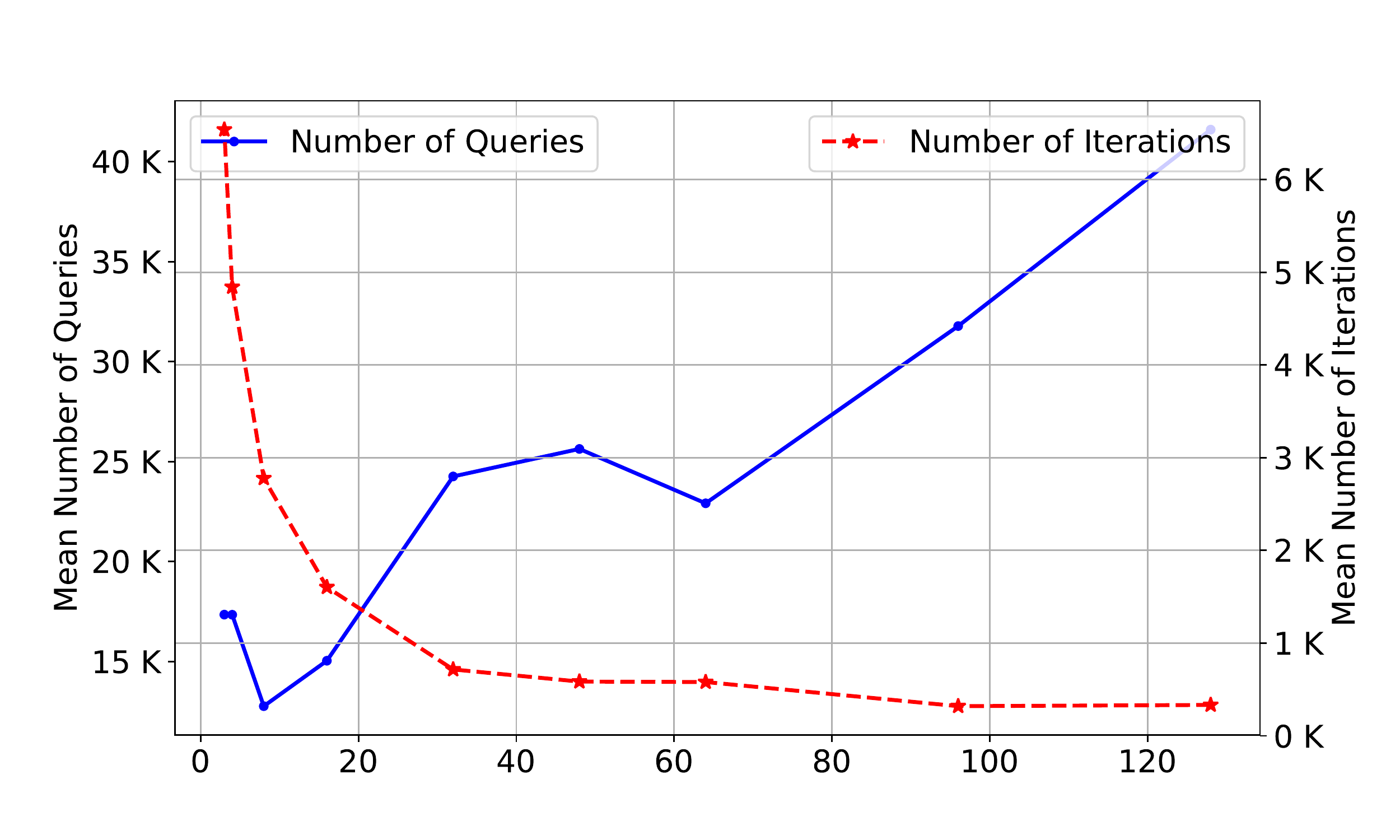}
    \caption{Effect of population size selection on both the speed of convergence and the number of queries.}
    \label{fig:query_pop_size}
\end{figure}
\subsubsection*{Population size:}
Large population size allows for increased population diversity and thus improved exploration of the search space within fewer iterations. However, since the evaluation of each population member costs one query, there is a trade-off in the selection of large population size to accelerate the algorithm success (in terms of the number of iterations), and the total number of queries spent. Figure~\ref{fig:query_pop_size} demonstrates this trade-off. On a set of $20$ images, we recorded the mean number of queries as well as the number of iterations until success under different choices of population size. From this experiment, we conclude that the relatively small population size of six is a reasonable choice to balance between convergence speed and number of queries.

\subsubsection*{Mutation rate:}
\begin{figure}[!t]
    \centering
    \includegraphics[width=0.9\columnwidth]{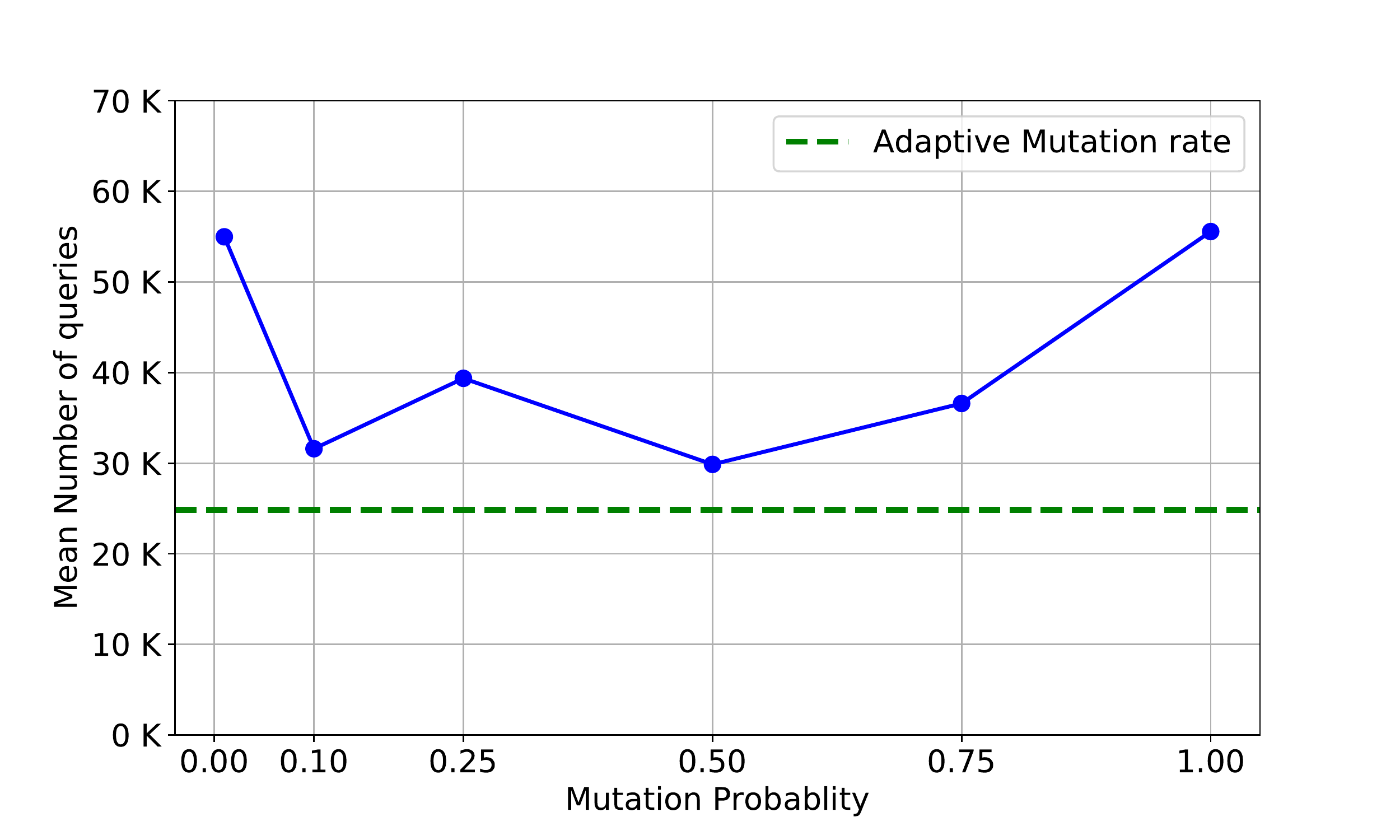}
    \caption{Effect of mutation probability on the number of queries required until success.}
    \label{fig:mutation_rate_effect}
\end{figure}

For the mutation rate  $\rho$, we found that the best result can be achieved if we use an adaptive mutation rate which is gradually decayed according to Eq.~\eqref{eq:decay_pho} and~\eqref{eq:decay_alpha} in Section~\ref{sec:algorithm}. As shown in Figure~\ref{fig:mutation_rate_effect}, this method performs better than fixed valued mutation rates. Effectively, the adaptive mutation rate balances between exploration and exploitation by encouraging exploration initially, and then gradually increasing the exploitation rate as the algorithm approaches convergence near an optimal solution.
\section{Conclusion}
\label{sec:conc}
\genattack is a powerful and efficient black-box attack which uses a gradient-free optimization scheme via adopting a population-based approach through genetic algorithms. We evaluated \genattack by attacking well-trained MNIST, CIFAR-10, and ImageNet models, and found that \genattack is successful at performing targeted black-box attacks against these models with not only significantly less queries than the previous state-of-the-art, but additionally can achieve a high success rate on ImageNet, which previous approaches are incapable of scaling to. Furthermore, we demonstrate that \genattack can succeed against ensemble adversarial training, the state-of-the-art ImageNet defense, with only a limited increase in the number of queries. Finally, we showed that \genattack can succeed against gradient obfuscation, due to its gradient-free nature, namely through evaluating against non-differentiable input transformations,  and can even succeed against randomized ones by generalizing the fitness function to compute an expectation over the transformation. To the best of our knowledge, this is the first demonstration of a black-box attack which can succeed against these state-of-the-art defenses. Our results suggest that population-based optimization opens up a promising research direction into effective gradient-free black-box attacks.

\begin{acks}
This research was supported in part by  the U.S. Army Research Laboratory and the UK Ministry of Defence under Agreement Number W911NF-16-3-0001, the National Science Foundation under awards \# CNS-1705135, IIS-1719097,  and OAC-1640813, and the NIH Center of Excellence for Mobile Sensor Data-to-Knowledge (MD2K) under award 1-U54EB020404-01. Any findings in this material are those of the author(s) and do not reflect the views of any of the above funding agencies. The U.S. and U.K. Governments are authorized to reproduce and distribute reprints for Government purposes notwithstanding any copyright notation hereon. 
\end{acks}

\bibliographystyle{ACM-Reference-Format}
\bibliography{references}

\end{document}